\documentclass[twoside]{article}

\usepackage{PRIMEarxiv}

\usepackage[utf8]{inputenc} 
\usepackage[T1]{fontenc}    
\usepackage{hyperref}       
\usepackage{url}            
\usepackage{booktabs}       
\usepackage{amsfonts}       
\usepackage{nicefrac}       
\usepackage{microtype}      
\usepackage{lipsum}
\usepackage{fancyhdr}       
\usepackage{graphicx}       
\graphicspath{{media/}}     
\fancypagestyle{plain}{%
  \fancyhf{}%
  \fancyfoot[RO,LE]{\thepage}%
}

\title{Towards complete digital twins in cultural heritage with ART3mis 3D artifacts annotator
}

\author{
  D. Karamatskos, V. Arampatzakis, V. Sevetlidis, G. Pavlidis \\
  Athena Research Center, Institute for Language and Speech Processing (ILSP) \\
  Xanthi, Greece \\
  \texttt{\{dkaramatskos, vasilis.arampatzakis, vasiseve, gpavlid\}@athenarc.gr} \\
  \And
  S. Nousias, A. Kalogeras, C. Koulamas, A. Lalos \\
  Athena Research Center, Industrial Systems Institute (ISI) \\
  Patras, Greece \\
  \texttt{\{nousias, kalogeras, koulamas, lalos\}@isi.gr} \\
}

\begin{document}
\maketitle
\thispagestyle{plain}

\begin{abstract}
Archaeologists, as well as specialists and practitioners in cultural heritage, require applications with additional functions, such as the annotation and attachment of metadata to specific regions of the 3D digital artifacts, to go beyond the simplistic three-dimensional (3D) visualization. Different strategies addressed this issue, most of which are excellent in their particular area of application, but their capacity is limited to their design's purpose; they lack generalization and interoperability. This paper introduces ART3mis, a general-purpose, user-friendly, feature-rich, interactive web-based textual annotation tool for 3D objects. Moreover, it enables the communication, distribution, and reuse of information as it complies with the W3C Web Annotation Data Model. It is primarily designed to help cultural heritage conservators, restorers, and curators who lack technical expertise in 3D imaging and graphics, handle, segment, and annotate 3D digital replicas of artifacts with ease.
\end{abstract}


\section{Introduction}

Since 1964, when ICOMOS laid out the basis for the preservation of monuments, cultural heritage has grown as a concept to include a wide range of tangible and intangible assets \cite{de2013values,unesco2003convention,unesco1978recommendation}. Soon, it was realized that curative approaches were not efficient and preventive strategies and policies should be adopted within the framework of a changing society, environment, and economy, leading to the 2009 UNESCO Chair on Preventive Conservation, Monitoring and Maintenance of Monuments and Sites. At the same time, the Industry 4.0 approach and the ongoing digitization in virtually every sector led to the development of the \textit{digital twin} concept at the dawn of the 21st century. The concept's origins can be traced to 2002 when Grieves presented a project for a Product Lifecycle Management Center \cite{grieves2002conceptual}, in which the basic principles were presented. Some 12 years later, Grieves defined the digital twin as a "virtual representation of what has been produced". Furthermore, it was highlighted that a digital twin might "(allow) to understand better what was produced versus what was designed, tightening the loop between design and execution" \cite{grieves2015digital,grieves2017digital}. 

The concept of digital twins also evolved outside the industry domain to represent any physical object or process and its historicity, the evolving digital profile of the status, interactions, and behaviors of the physical object or process through space and time. The evolution of the networking and technological infrastructures and the pervasive IoT applications widened even the application domain, leading to the consideration of a technological framework applicable to cultural heritage. This framework is envisioned as a highly multi-layered ontology on top of a digital replica of heritage assets (see, for example, \cite{klein2017wireless}). To apply and implement digital twin approaches for the protection of cultural heritage requires a cross-disciplinary collaboration, including cultural heritage researchers, conservation experts, engineers, and information technology specialists \cite{letellier2007recording}. 

Towards the complete adaptation of the digital twin concepts in cultural heritage, archaeology, and cultural heritage applications increasingly use 3D artifacts, sensor data, and computational methods. More advanced features than those provided by the baseline 3D visualization are becoming essential. The capability to attach multi-layered representations as metadata on top of a cultural object's 3D replica is among these requirements. These approaches need to be supplemented or combined with what is known as \textit{annotation functions} in order to be able to provide such rich settings for visualization and further analysis, including artificial intelligence applications for knowledge extraction.

The process of choosing a region on a 3D surface and connecting that region to a high-level representation, either in structured or free form, is known as \textit{3D object annotation}. Two types of annotation on 3D objects are commonly considered, (a) automated feature or semantics-based region segmentation \cite{nousias2020saliency}, (b) manually, or, in some cases, computer-assisted, selected region characterization \cite{arnaoutoglou20033d}. While the former relates to automated semantics extraction for various computer vision and graphics applications, the latter is most suited to cultural heritage applications, like conservation. In this scenario, 3D artifacts are presented to experts, which are called to annotate either the various degradation effects (humidity damage, surface cracks, and fissures) or any restoration processes that took place (preservation information, mended areas, cleaning process) \cite{ponchio2020effective,soler2013design}. Moreover, materials and information correlated to decoration can also be annotated.

Not long ago, implementing an annotation tool did not follow any guidelines. No standardization existed; each research team and project devised a new method, idea, and approach that solved targeted or domain-specific challenges. Currently, W3C has released a standard regarding  web annotations\footnote{See \url{https://www.w3.org/TR/annotation-model/}.}.

Among the most challenging issues in implementing a 3D annotation system are (a) the user interaction design, (b) the input conversion and association, (c) the management of high-resolution 3D data, (d) the representation schemes, (e) the annotation format and data handling, and (f) the annotation as a Web application \cite{ponchio2020effective}.

The motivation behind creating the \textit{ART3mis} annotation tool was to address the issues mentioned above and give heritage conservators a user-friendly, interactive, web-based textual annotation tool for 3D cultural artifacts. The idea originated in the EU project WARMEST\footnote{Project WARMEST website, \url{https://warmestproject.eu}}, where an annotation tool was needed for the annotation of the degradation on the 3D models of the columns in the \textit{Patio de los Leones}, in the UNESCO World Heritage Site of Alhambra, Granada, Spain. As the tool was required for scientific use by experts with no technical skills in 3D imaging and graphics, ART3mis had to follow the intuitive What You See Is What You Get (WYSIWYG) design. This approach enables users to handle quickly and in real-time the various 3D models created by the 3D object digitization of the monument. Image-based digitization was employed in WARMEST, and highly detailed 3D replicas were created \cite{pavlidis2007methods}; thus, models of high polygon counts needed to be handled in real-time. Although ART3mis was primarily tested with the WARMEST 3D models, it was also tested with other 3D models with geometry in the range of 20 million polygons. ART3mis was presented at the CAA 2021 International Conference \cite{arampatzakis2021artemis}. It was initially a Windows-only executable desktop application based on the geometry processing library known as \textit{libigl}. ART3mis was created due to the absence of an accessible tool for complex area annotation.

This paper presents the significantly upgraded and most complete version of ART3mis. The current version, virtually a completely new software, features a Web application that supports manual and automated annotation, improved graphics performance, portability, intuitive ROI selection methods, and complex structured annotations for enhanced semantics support. ART3mis was developed and is presented in one of the most challenging frameworks, the annotation of 3D digitized cultural objects, which are highly complex, irregular, and generally non-watertight digital entities, coming in large data volumes and various formats. The presentation that follows consists of a brief description of the background and the related work regarding 3D object annotation (section 2), a presentation of ART3mis, its features, and the supported functionalities (section 3), concluding with a discussion on the work and future perspectives (section 4).

\section{Background and related work}
 
In 3D object annotation, based on the data types, annotations can be attached to single points, lines, voxels, or regions on a 3D object. The following types of annotations are defined: 
\begin{itemize}
\item 
\textbf{Surface space}: On the surface of the objects (3D mesh), as collections of polygons or points.
\item 
\textbf{Texture space}: On the texture of the objects, in UV-space, apparently leading to a 2D annotation approach.
\item 
\textbf{Hybrid 2D-3D projected image space}: On a hybrid space, fusing annotation on 2D images with reprojection techniques to transfer the annotations in 3D space.
\item 
\textbf{Voxel Space}: Collections of voxels.
\end{itemize}
For a detailed presentation of the various approaches, the interested reader may refer to \cite{ponchio2020effective}. 

In a nutshell, 3D object annotation consists of defining Regions of Interest (ROIs) in a segmentation manner and attaching specific data or information to them. Focusing on 2.5D objects (3D objects defined only by surfaces), two types of ROI selection are possible, segmentation and trimming; trimming entails changing the polygons to fit the user-selected region, whereas segmentation involves choosing a collection of polygons from the original 3D model. In any case, a trimmed region is a subset of the corresponding segmented region. Trimming necessitates more graphics processes yet may support a more precise region selection. Additionally, the procedure modifies the 3D mesh to fit the chosen ROI, making it an intervention technique. An inherent challenge in 3D object annotation is the correct selection of 3D surfaces while being viewed on a flat 2D screen, where the 3D scene is a projection onto the screen's surface. The distances of the surfaces from the viewer (what is called \textit{depth}) are unknown (the third dimension is lost). Typically, in 3D annotation applications, the desired ROI is expected to be visible (rendered); thus, there must be a mechanism to define the depth based on the rendered surfaces. Another highly relevant challenge is to select only the part of the 3D object that is of interest and not everything that happens to be projected within the same selected region (potentially hidden behind the foremost surfaces). Another challenge in creating a 3D mesh-based annotation tool is in providing the ability to select ROIs of any form with friendly and intuitive interaction techniques while relying on data interoperability standards. These challenges will be discussed later in the presentation of ART3mis. Regardless, 3D object annotation is a process of high utility for various domains; the relevant scientific literature is not equally extended. Comprehensive approaches toward 3D object annotation are as follows: 
\begin{itemize}
\item
\textit{ShapeAnnotator} \cite{attene2009characterization} is a tool that matches a 3D object with instances of ontology classes, utilizing multiple automatic segmentation algorithms and providing an ontology browser. The user is provided the means to edit the ROIs, and the results are saved in an OWL file.
\item
\textit{3D-COFORM} \cite{serna2012interactive} is the first interactive semantic enrichment tool based on the CIDOC-CRM ontology. The region selection is performed using 3D primitives, such as spheres or cylinders, and by outlining contours on the surface of the 3D model. Furthermore, 3D-COFORM has been designed so that multiple users can work on the same 3D model using a shared repository.
\item
The \textit{3D Semantic Association (3DSA)} \cite{yu2013documenting} is a Web system that enables the analysis of 3D cultural objects based on specific ontologies and supporting annotation, which may also be semantically assisted. ROIs are selected based on user-defined polygonal shapes or 3D volumetric segments overlaid on the rendered model. The underlying ontology in the system provides the potential to support high-level semantics and inferences.
\item
\textit{POTREE} \cite{schutz2016potree} is a Web viewer for 3D point clouds equipped with several measurement tools for large point clouds. POTREE seems to support only point-based annotation, with descriptions that can be edited in an HTML file. POTREE has the potential to be used as the base engine for the development of a more focused Web-based annotation application.
\item
\textit{CHER-Ob} \cite{shi2016cher} is a heritage object analysis tool that also supports 3D mesh annotations. Annotations are based on rectangular region selections and can either correspond to surface patches or parts of the 3D model. It should be noted that CHER-Ob is a complete content management system with additional analysis tools for artifacts.
\item
\textit{ClippingVolumes} \cite{ponchio2020effective} is an advanced annotation approach, supporting arbitrary shaped user-defined region selections. This method's innovation is in determining the correct surface to be selected. The ROI selection is addressed by defining an arbitrary bounding polyhedron, which is minimized iteratively. A content management system supports the storage of the annotations.
\end{itemize}

\section{ART3mis 2.0}

As explained in the Introduction, ART3mis was initially conceived as a user-friendly annotation tool for 3D digitized cultural heritage objects (in particular parts of historical buildings, like columns) in the framework of the EU project WARMEST. Its purpose was to provide conservation scientists and specialists with a tool to easily attach information on 3D surfaces regarding the degradation of materials. The annotated objects would serve both as digital twins for the original objects and as the ground-truth dataset for developing machine learning techniques capable of detecting defects or degradation in cultural digital twins. The requirements were highly challenging, including (a) the handling of various 3D file types, (b) the handling of 3D digitized objects, usually of several GBytes in size, (c) the provision of WYSIWYG interaction, (d) the compliance with international standards in data exchange. Further, in the new, totally reworked version or ART3mis, more requirements were added, including (a) the Web functionality, with more security challenges, (b) the real-time graphics functionality, (c) the support of more complex structures for the annotations and (d) the connection with automated annotation modules.

ART3mis can provide several ways of ROI selection on 3D model surfaces. It operates \textit{on the textured 3D model}, which has been widely discussed as the most appropriate approach in the relevant literature \cite{ponchio2020effective}. Methods using texture accurately detect interest points on various materials. ART3mis provides a complete annotation management tool that allows annotating an object for the first time or editing existing ones. In order to allow the distribution and reuse of said annotations across various hardware and software platforms, the annotations are expressed via a JSON-LD data structure compliant with the W3C Web Annotation Data Model. This approach is increasingly used in modern Web-connected applications and is regarded as the industry standard by W3C. ART3mis adheres to the WYSIWYG philosophy and the ten heuristic criteria that describe a user-friendly interface \cite{nielsen1990heuristic} because its target audience consisted of heritage conservators who, potentially, lacked technical expertise in 3D imaging and graphics:
\begin{enumerate}
\item	Visibility of system status
\item	Match the system and the real world
\item	User control freedom
\item	Consistency and standards
\item	Error prevention
\item	Recognition rather than recall
\item	Flexibility and efficiency of use
\item	Aesthetic and minimalist design
\item	Help users recognize, diagnose, and recover from an error
\item	Help and documentation
\end{enumerate}

\begin{figure}[t!]
\centering
\includegraphics[width=.75\linewidth]{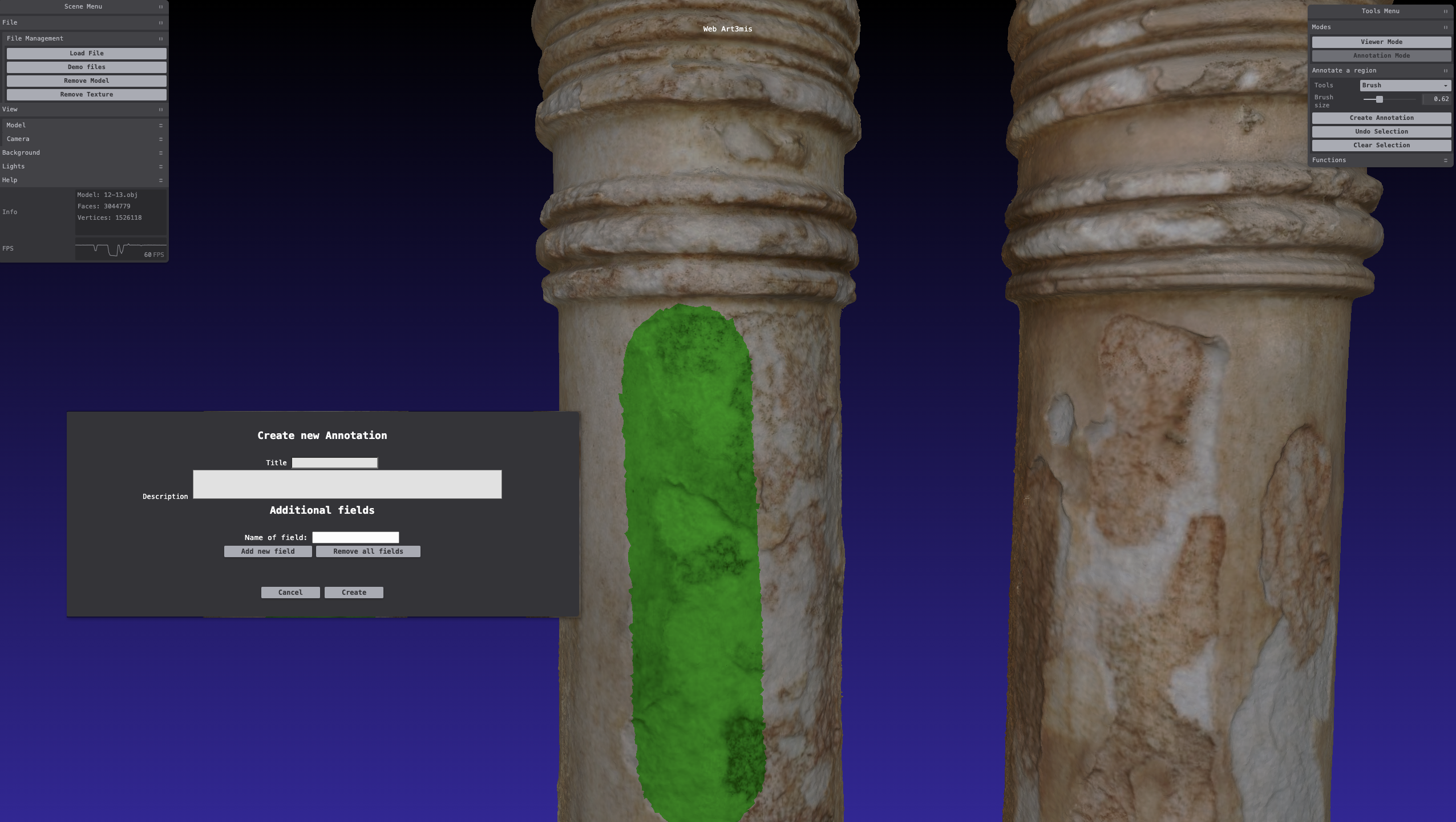}
\caption{Brush selection in ART3mis 2.0.}
\label{fig1}
\end{figure}

\begin{figure}[t!]
\centering
\includegraphics[width=.75\linewidth]{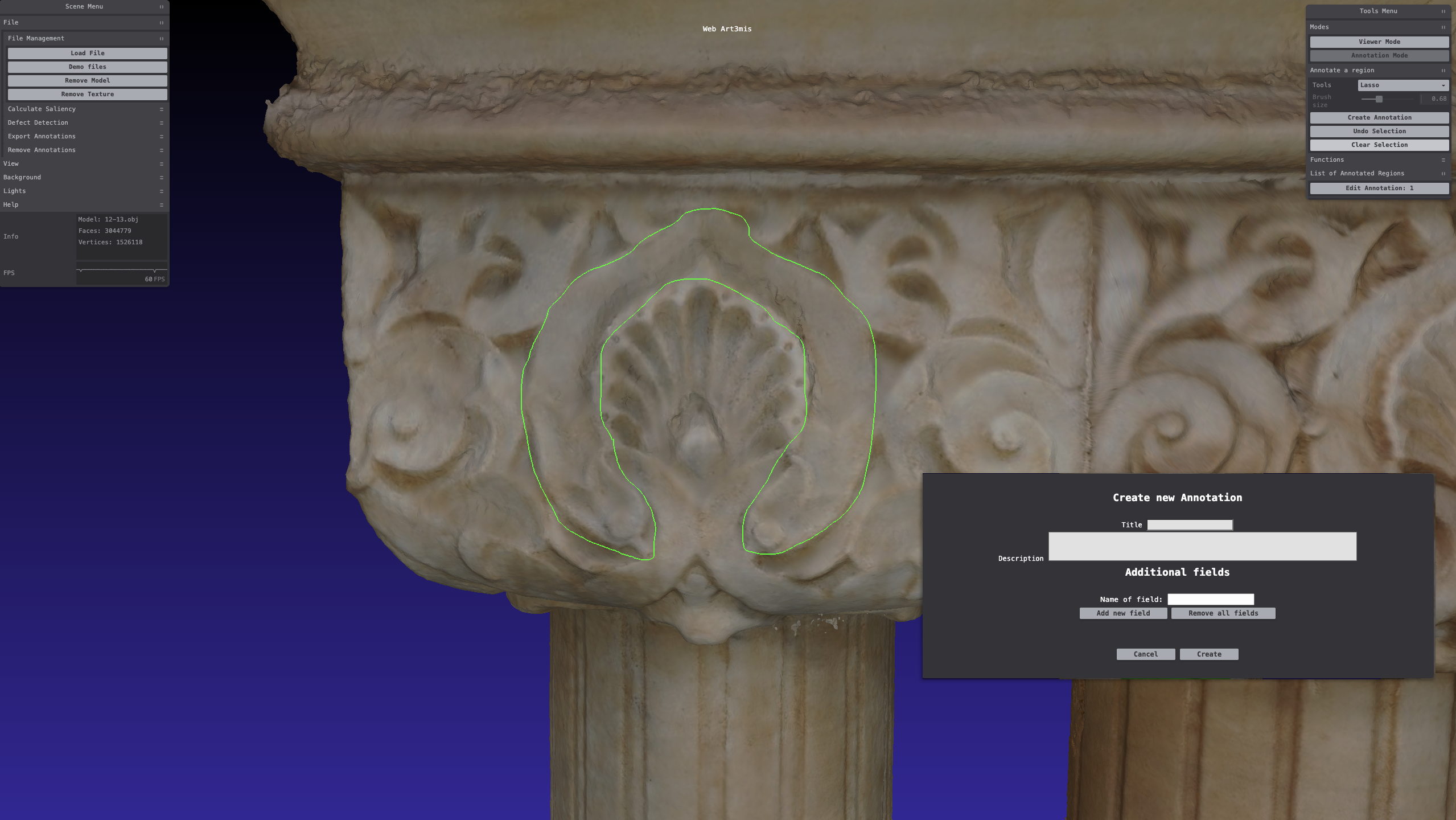}
\caption{Lasso selection in ART3mis 2.0.}
\label{fig2}
\end{figure}

Regarding \textit{manual ROI selection}, ART3mis supports two methods: (a) brush/painting selection and (b) polyline/lasso selection. It is emphasized that in either of the two modes of operation, the selection of ROIs is based on the selection of 3D polygons that correspond to the parts of the 3D mesh bound or defined by the user-selected region using \textit{ray intersection}. In computer graphics, a ray-polygon intersection is finding any or all points that simultaneously lie on a particular ray, and a particular polygon \cite{glassner1989introduction}. Thus it is a mature and successful approach to selecting parts of a 3D mesh that are visible (rendered) on display. From a technical perspective, user selections via input devices, such as mouse controls,  cast a ray from the pointer's position through the frustum of the scene towards the object; any intersections have occurred, paint the regions on the 3D model surface correspondingly. In \textit{brush selection}, a typical brush tool is used to paint on the screen, like any brush in most applications. The width of the brush is user-defined to allow for control of the speed and accuracy of the selection. The function of the brush tool is showcased in \figurename~\ref{fig1}. In \textit{lasso selection}, the user may draw a free-form shape enclosing an ROI on the visible (rendered) 3D object, like any other lasso tool in other applications. Ray-polygon intersection also underlies the selection functionality in this case, which is showcased in \figurename~\ref{fig2}.

Regarding the \textit{meta-information storage}, each annotation is split into two parts as follows: (a) the central part and (b) the metadata part. The first contains the faces, vertices, and the corresponding description. The latter contains the metadata part that holds a user-defined ROI title, a preferred display color, and a unique ID of the ROI. A unique feature of ART3mis 2.0 is the support for complex data structures on each ROI.


Regarding \textit{automated ROI selection}, a feature unique to ART3mis 2.0, the tool supports the connectivity with Web service modules that interchange 3D model data and respond with regions of interest based on particular algorithms that detect various user-dependent interest points or regions. In the current implementation of ART3mis, two Web modules have been connected with the system, supporting the detection of defects and saliency maps, as will be explained in the following sections.

\subsection{ART3mis 2.0 implementation specifics}

\begin{figure}[t!]
\includegraphics[width=\linewidth]{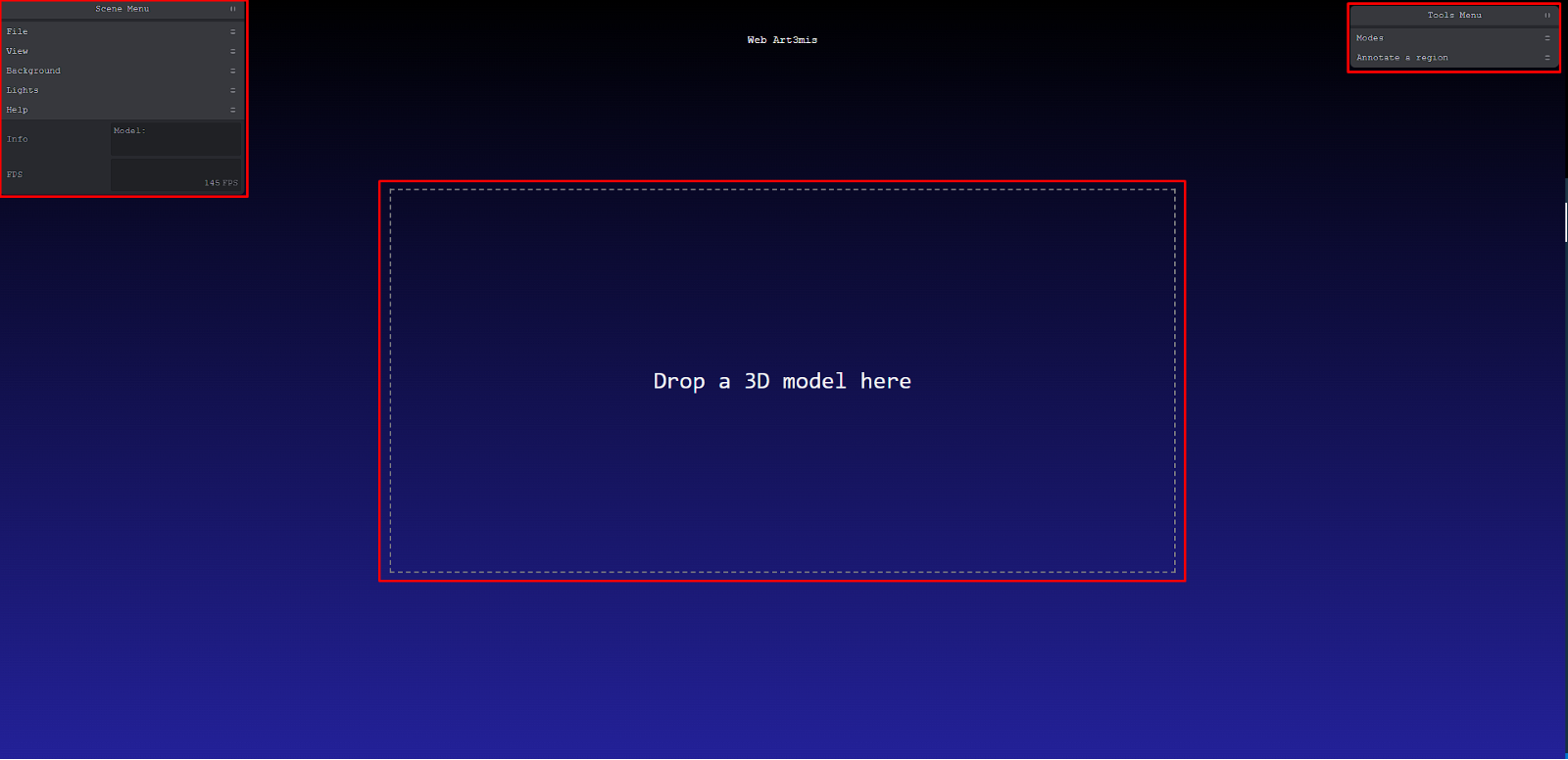}
\caption{ART3mis 2.0 Web user interface.}
\label{fig4}
\end{figure}

ART3mis 2.0 is a client-side browser application. It is based on a cross-browser Java\-Script library and application programming interface called Three.js. The latter's function is to create and display animated 3D computer graphics in a web browser through the WebGL API. Furthermore, Tweakpane is responsible for the intuitive yet straightforward interface. Moreover, ART3mis supports many standard 3D object file formats from various 3D imaging and computer graphic platforms, such as Blender and MeshLab, as well as it supports 3D models from 3D digitization projects. The annotations are saved in a JSON-LD data structure. Unlike relevant applications, the ART3mis development approach focused on balancing a user-friendly design and a fully featured annotation application. To this end, the application presents three main interaction frames, as described below and shown in \figurename~\ref{fig4}:

\begin{itemize}
\item 
\textbf{Scene menu} (top-left): menus for file loading or export, settings, model and rendering information, and help.
\item 
\textbf{Tools menu} (top-right): menus for the annotations, annotation mode selection, and selection tool settings.
\item 
\textbf{Workspace Area} (middle): main region for the view and manipulation of 3D models and annotations.
\end{itemize}
 
ART3mis provides flexible user-friendly and intuitive tools, which also support fast and accurate annotation, in contrast with similar annotation systems, in which the users select a rectangular region or a volume of a 3D model \cite{shi2016cher}.

\subsection{Manual annotation}

Loading a 3D model and any accompanying data, such as texture, metadata, or saved annotations, is the first step in the manual annotation process in ART3mis. The loaded 3D model is manipulated using standard mouse-keyboard commands that support rotation, panning, and zooming, like in any other 3D object manipulation tool. As described in previous paragraphs, two manual region selection modes are supported and operated by simple mouse interactions. A user selects an ROI, associates an annotation, then the results are structured as an XML and subsequently stored at the body of the W3C Web Annotation Data Model. The annotation structure is constantly updated during user interactions. Annotation editing is also supported for an arbitrary number of annotations per 3D model, including potentially overlapping annotations. A unique feature of ART3mis 2.0 is that, the user can define a custom "database" and end up with a complex data structure for each ROI.

\subsection{Automated annotation}

The automatic ROI selection in ART3mis 2.0 is modular. It can employ various detection algorithms in the form of Web services to tailor to the needs of various applications. Experts from different domains can develop bespoke detection algorithms and connect them with the proposed software. This unique feature of ART3mis is currently exploited to produce fast and accurate ROI identification and selection using advanced machine learning algorithms. Subsequently, the annotation of the automatically selected ROIs follows the same workflow as the manual annotation, resulting in JSON data structures. The automated and manual annotation workflows can be used concurrently, complementary, and interchangeably, resulting in fast and accurate complex annotations on 3D objects. Currently, ART3mis offers two automatic selection methods based on (a) saliency detection and (b) defect detection, which have been separately published and are briefly described in subsequent paragraphs. In essence, the former select regions on the 3D model characterized by high saliency, whereas the latter, as the name implies, "semantically" selects regions that correspond to patterns of potential surface or material defects, like corrosion and rust. 

\begin{figure}[t!]
\centering
\includegraphics[width=.5\linewidth]{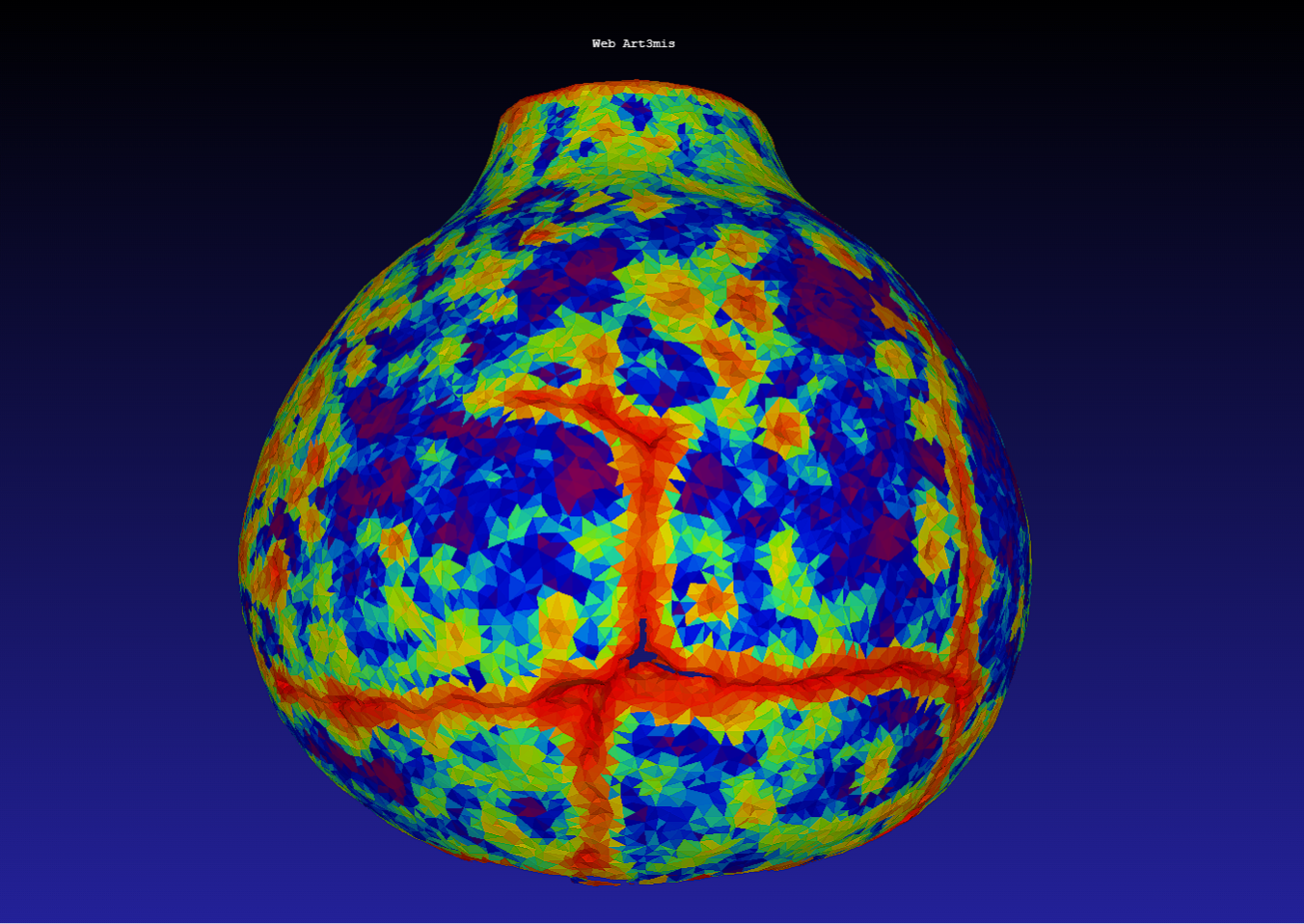}
\caption{Automated saliency detection in ART3mis 2.0.}
\label{fig7}
\end{figure}

Geometric details of a 3D model, such as high-frequency features, and surface areas consisting of points of interest (e.g., cracks in an ancient vase), are considered salient. Thus they must be handled differently from the rest of the 3D model (i.e., smooth or flat areas), especially in applications that deal with sensitive content, such as the 3D representation of historical objects. In this application, any perceptual detail or possible "geometry divergence" of the object must be preserved since they represent valuable information that can be used through the object's maintenance and preservation process. ART3mis's saliency detection algorithm uses a CNN-based saliency map extraction pipeline that annotates the most critical information of a 3D model. These include geometric details such as the delicate features of the model or surface defects \cite{nousias2020saliency}. The outcome of the saliency detection algorithm is fed back to the ART3mis annotator, so the users can observe it as a heat map, as shown in \figurename~\ref{fig7}. In this map, the "hotter" the area, the higher the importance of the corresponding region. This method produces a colorful representation of various regions that may be used as focal points or points of interest for further inspection and annotation.

\begin{figure}[t!]
\centering
\includegraphics[width=.5\linewidth]{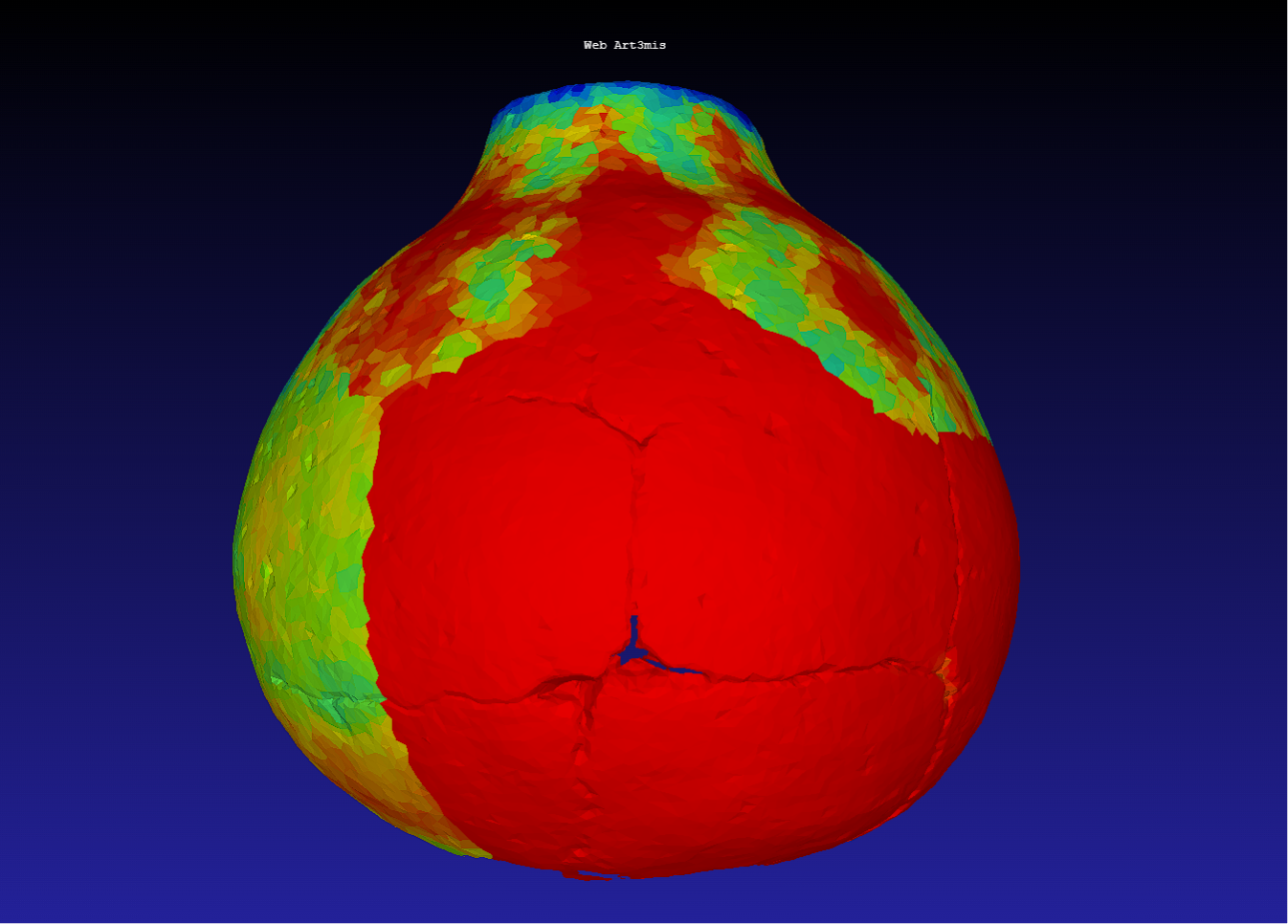}
\caption{Automated defects detection in ART3mis 2.0.}
\label{fig6}
\end{figure}

Furthermore, 3D model analysis techniques have been proposed to detect defects on 3D objects as an alternative to on-site inspections. ART3mis's defect detection algorithm uses a Convolutional Neural Network (CNN) for the automatic annotation on the surface of the 3D model, which highlights areas potentially corresponding to surface material defects \cite{arvanitis2022coarse}. As shown in \figurename~\ref{fig6}, a heat-map indicates the susceptibility of the surface to defects, like corrosion, rust, and the like. As in the previous case, this method produces a colorful representation of various regions that may be used as focus points for further inspection and annotation.

The web-based aspect of ART3mis 2.0 allows users to access the application from any computer with internet access, regardless of their operating system. A user can manage multiple complex annotations with the supported JSON-LD structures. Last, ART3mis can generate intuitive reports of created annotations in PDF files, which can be shared among colleagues, researchers, and interested parties.

\section{Conclusion}

ART3mis is a brand-new 3D model, web-based annotation tool that provides user-friendly, real-time, multiple JSON-encoded simple text annotations on high-resolution 3D models. It adheres to the ten heuristic criteria that define a user-friendly environment and the WYSIWYG concept. A unique ray intersection and selection volume approach forms the foundation of the ART3mis region selection engine, which guarantees precise real-time 3D mesh selection. Support for section cuts, tomography operations, and mobile implementation can all be added to ART3mis future versions.

ART3mis is available online at \url{http://art3mis.athenarc.gr/}.


\section{Acknowledgments}
This work was supported by European Union Horizon 2020 Research and innovation program ``WARMEST-loW Altitude Remote sensing for the Monitoring of the state of cultural hEritage Sites: building an inTegrated model for maintenance'' under Marie Sklodowska Curie grant agreement No 777981.

\bibliographystyle{unsrt}  
\bibliography{references}

\end{document}